\documentclass[journal]{IEEEtran}
\usepackage{booktabs}
\usepackage{graphicx}
\usepackage{amsmath}
\usepackage{subcaption}

\usepackage{tikz}
\usetikzlibrary{positioning,arrows.meta,calc}

\ifCLASSINFOpdf
\else
\fi
\usepackage{url}

\begin{document}
%

 \title{
  Admittance-Based Surface Alignment for Human-in-the-Loop Robotic Visual Inspection
}
%
%
%

\author{Antara~Banerjee*,
        Colin~Acton*,
        and~Xu~Chen
\thanks{Antara Banerjee is with the Department
of Electrical and Computer Engineering, and Colin Acton and Xu Chen are with the Department of Mechanical Engineering, University of Washington, Seattle. * denotes equal contribution.}
\thanks{\\Website: \protect\url{https://macs-lab.github.io/robotic-inspection-orientation-control-2026/}}}

\maketitle

\begin{abstract}
Precision visual inspection underpins quality assurance across aerospace, semiconductor, and medical manufacturing, where undetected surface anomalies on high-value parts translate directly into scrap, rework, and field failures. Robotic visual inspection requires precise alignment between the end-effector and local surface geometry in the presence of perception noise and surface irregularities. In industrial settings, a human operator is often kept in the loop via teleoperation or shared autonomy, introducing real-time adjustments that render purely offline motion planning inadequate. This motivates control architectures capable of reactive, compliant behavior under combined human and perceptual uncertainty. This paper presents a novel real-time, closed-loop robotic orientation control pipeline for precision visual inspection, with an admittance-based framework that unifies operator input and perception-driven surface alignment. We design the end-effector as a virtual sphere moving through a viscous medium, such that the resulting physically interpretable mass--damper system generates synchronized, compliant motion from orientation error and operator commands. We validate the framework on a 6-DOF manipulator demonstrating stable normal-tracking and a final mean orientation error of $0.4^\circ$.

\end{abstract}

\begin{IEEEkeywords}
Admittance control, robotic inspection, shared autonomy, compliant control, surface normal tracking, human–robot interaction
\end{IEEEkeywords}

%
\IEEEpeerreviewmaketitle

\section{Introduction}
\label{sec:intro}

Modern advanced manufacturing, spanning aerospace composites, semiconductor packaging, medical devices, and additively manufactured components, depends on high-throughput visual inspection to enforce sub-millimeter geometric and surface-quality tolerances. Undetected surface anomalies on high-value parts propagate into costly rework, scrap, and downstream field failures, making in-line precision inspection a decisive determinant of yield, cost, and product safety. As geometries grow more complex and production volumes scale, manual inspection alone can no longer meet the speed, repeatability, and accuracy demands of these industries, motivating the broad shift toward robot-mounted imaging systems for routine, high-precision surface assessment.

Automated inspection systems consisting of eye-in-hand manipulators running coverage path planning \cite{Nandagopal2024} and anomaly detection models are experiencing increased prominence and academic investigation as product complexity and production volumes increase~\cite{mullany}; however, the need for human inspectors to interface with these systems has largely been ignored. Automated systems are increasingly adept at the part recognition, search, and detection stages of visual inspection, while humans are still essential for proper classification and action tasks~\cite{wang}.

\begin{figure}[t]
    \centering
    \fbox{\includegraphics[width=0.9\linewidth]{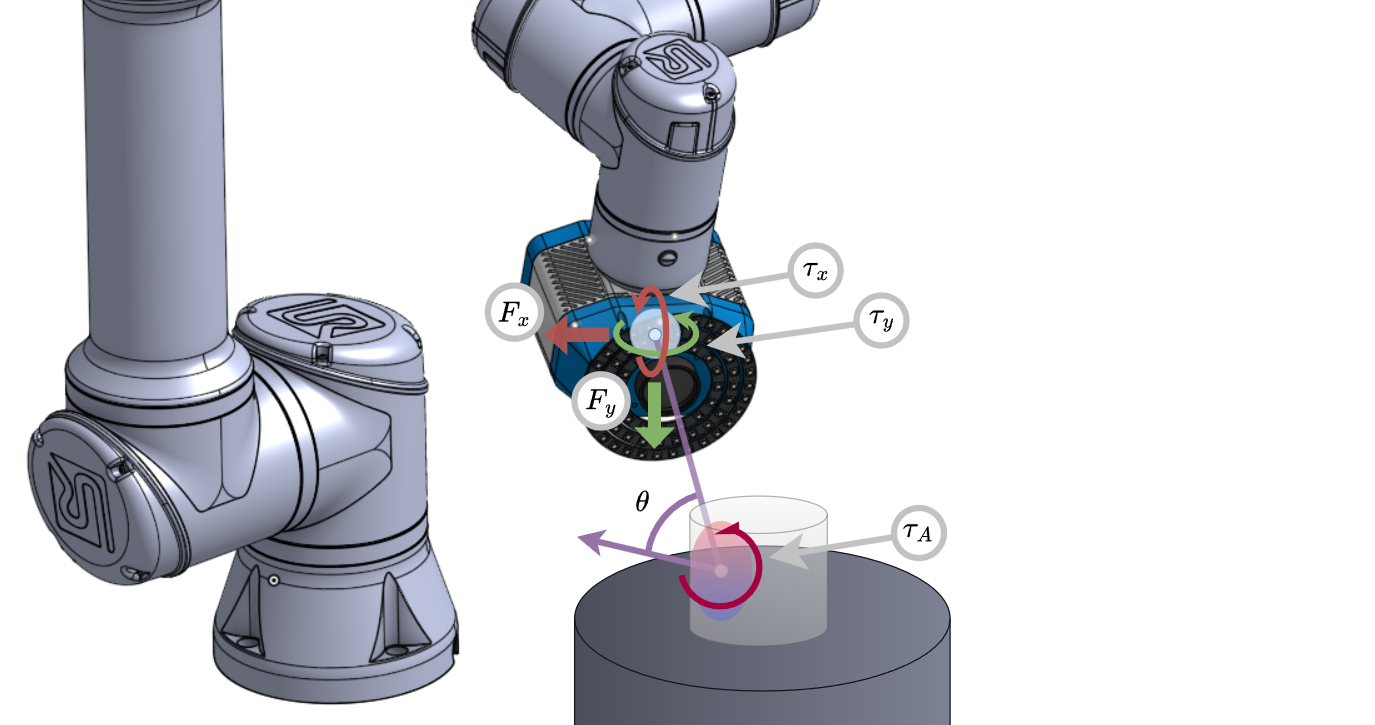}}
    \caption{Rendering of robotic inspection system. Real-time depth observations are used to calculate deviation from surface normal and produce simulated task space forces to realign the camera in concert with human operator inputs.}
    \label{fig:system_rendering}
\end{figure}

\begin{figure*}[t]
\centering
\includegraphics[width=\linewidth]{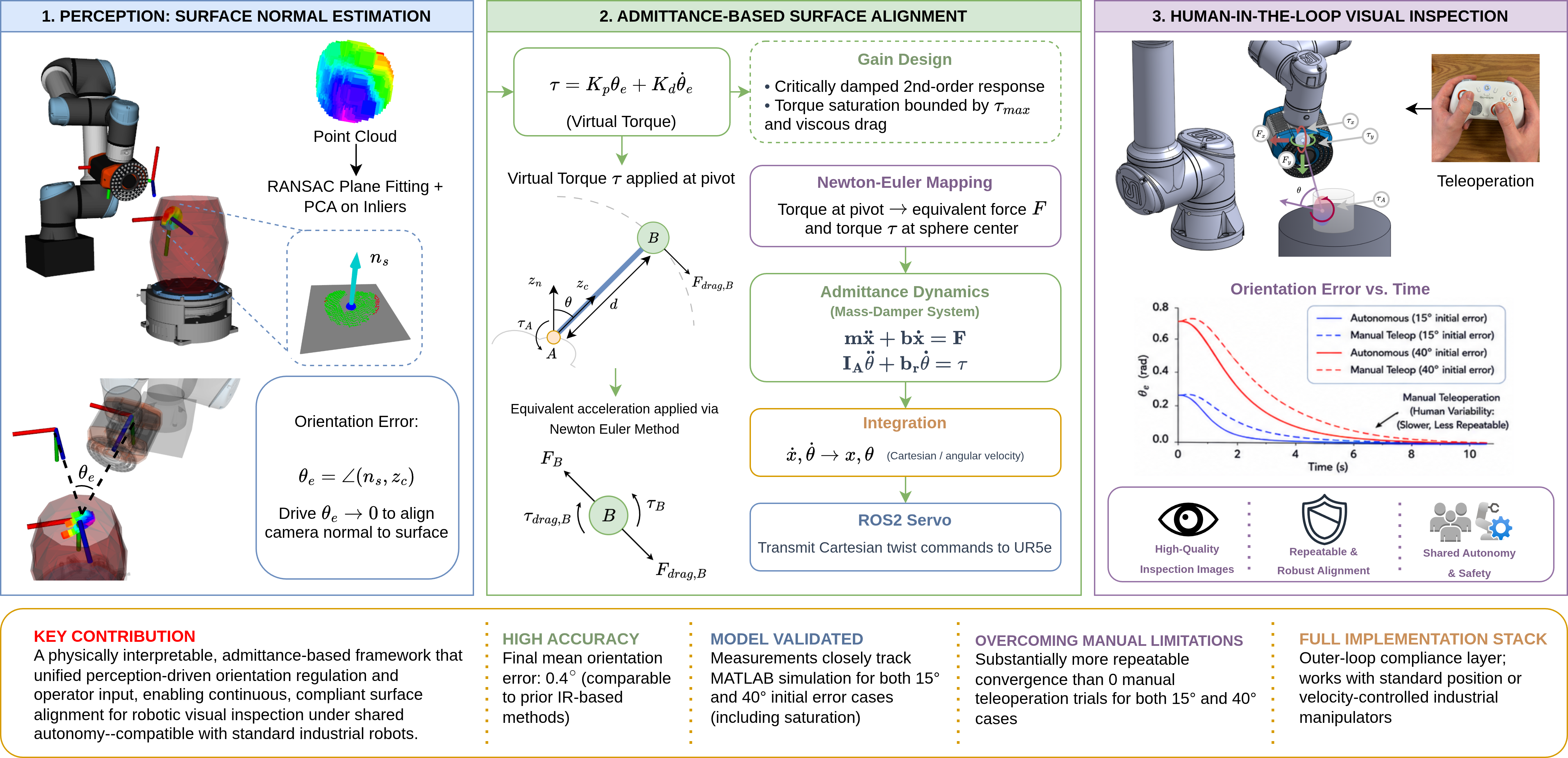}
\caption{Summary of the proposed approach to human-compliant, perception-based orientation control for inspection assistance.}
\label{fig:graphic_abstract}
\end{figure*}

When an anomaly is flagged, a human inspector will often need to follow up with an investigation to verify the model's accuracy and act upon it. In this stage, their ability to operate this precision equipment can become a bottleneck for production efficiency and a hurdle for adoption. To aid them in this task, high-level controllers can be designed to regulate the position and orientation of the end-of-arm imaging system, maintaining focus and normality with the surface of the object under inspection and freeing the inspector from 4 of the 6 degrees of operational complexity. Existing methods of orientation adaptation are designed without human compliance in mind, incorporating trajectory planning which prohibits human interaction. Meanwhile, methods incorporating real-time control loops utilize limited perception equipment, and do not react intuitively to inspector input.

In this paper, we present a first-of-its-kind application of real-time, closed-loop robotic orientation control to precision visual inspection, realized as a perception-driven orientation control aid for surface-adaptive, compliant human operation of a robotic visual inspection end-of-arm tool system. Fig.~\ref{fig:system_rendering} illustrates the physical configuration, in which an eye-in-hand depth camera streams real-time observations of the local surface to a controller that continuously realigns the end-effector with the estimated surface normal while remaining compliant to operator input. Fig.~\ref{fig:graphic_abstract} summarizes the overall approach, from depth-based normal estimation through the virtual mass--damper admittance layer to the velocity commands executed by the manipulator.
\noindent{The main contributions of this work include:}
\begin{enumerate}
\item A new perception-driven, closed-loop orientation regulator that couples real-time surface-normal estimation with a virtual mass--damper outer-loop compliance framework, such that the end-effector continuously realigns to local surface geometry without requiring pre-computed inspection trajectories.
\item An admittance-based control formulation that maps orientation error, teleoperation inputs, and human-in-the-loop corrections into a single compliant motion stream. This enables operator commands and perception-driven corrections to coexist while respecting actuator torque and velocity constraints.
\item We design the controller as a modular outer-loop architecture issuing task-space velocity commands through standard servo interfaces, such that the framework deploys on industrial position/velocity-controlled manipulators without modifying their internal control architecture.
\item We validate the framework experimentally on a UR5e platform with eye-in-hand depth sensing, demonstrating stable convergence, robustness to perception noise, and a final mean orientation error of $0.4^\circ$ that matches prior infrared-based methods while additionally supporting real-time, human-in-the-loop operation.
\end{enumerate}

\section{Related Works}
\label{sec:related}

Existing approaches to robotic surface following and visual inspection predominantly rely on offline planning strategies derived from depth sensing, particularly using RGB-D cameras to reconstruct object geometry and generate inspection trajectories. For instance, vision-guided inspection systems construct a global 3D model of the environment from multiple depth images and subsequently compute surface-following paths prior to execution \cite{nakhaeinia2016surface}. Similarly, recent works on automatic path planning for visual inspection generate trajectories directly from point clouds obtained from a depth camera, where surface normals and object geometry are used to define inspection poses in a pre-processing stage. These methods, while effective, largely operate in an open-loop or pre-planned manner, assuming that the reconstructed geometry sufficiently captures the environment and that execution conditions remain unchanged \cite{tasneem2023automatic}. Furthermore, the limited instances of online orientation regulation reported in the literature often rely on alternative sensing modalities such as infrared or proximity sensing, rather than directly leveraging real-time depth perception \cite{nakhaeinia2018hybrid}. While some methods incorporate in-process normal estimation and real-time pose correction, these approaches operate in a discrete perception–correction pipeline rather than a continuous closed-loop control framework \cite{yang2025adaptive}.

Human-robot collaborations, such as collaborative robotic assembly and inspection, require controllers that regulate the dynamic relationship between the motion of the end-effector and the interaction forces.
However, traditional industrial manipulators are primarily designed for precise position tracking and, therefore, struggle to perform tasks involving sustained physical interaction with uncertain environments. 
Early work on robotic interaction control focused on modifying robot motion based on measured forces during contact. Whitney introduced the concept of \emph{accommodation control}, where force feedback is used to modify the commanded motion of the manipulator in order to achieve stable contact during fine-motion tasks \cite{Whitney1977}. 
Raibert and Craig proposed the framework of \emph{hybrid position/force control}, which separates the control problem into motion-controlled and force-controlled directions \cite{RaibertCraig1981}. 
Hogan later formalized the concept of \emph{impedance control}, which regulates the dynamic relationship between force and motion by imposing a desired mechanical impedance at the robot end-effector \cite{Hogan1985}. In this framework, the robot behaves as a virtual mass–spring–damper system that determines how motion responds to applied forces, enabling compliant interaction with uncertain environments.

While impedance control provides a powerful dynamic interaction framework, many industrial robotic systems are position-controlled and do not provide direct torque control at the actuator level. As a result, interaction control is often implemented through outer-loop controllers that modify position or velocity commands based on measured forces. 
Position-based impedance control, also known as 
\emph{admittance control}, has therefore become a widely adopted approach for interaction control. In admittance control, measured interaction forces are mapped to desired motion commands through a virtual dynamic model, allowing compliant behavior to be implemented using position or velocity-controlled manipulators. Newman analyzed the stability and performance limits of interaction controllers \cite{Newman1992}.
Schimmels and Peshkin studied the robustness of an admittance control law for force-guided assembly and showed that force-to-motion mappings can be designed to remain effective even in the presence of contact friction disturbances \cite{Peshkin1992}.

Admittance control has been widely adopted in applications involving physical human--robot interaction and haptic interfaces, where the interaction force measured at the robot end-effector is used to generate motion commands through a virtual dynamic model. In this framework, a desired mass--spring--damper behavior is imposed such that the measured external force determines the commanded velocity or position of the robot. Keemink et al.~\cite{Keemink2018} provide a comprehensive overview of admittance control, clarifying its relationship with impedance control and reviewing its applications in haptic devices, rehabilitation systems, and collaborative robots. 
Duchaine and Gosselin proposed a velocity-based variable impedance framework that allows robots to adjust their interaction behavior during cooperative manipulation tasks \cite{Duchaine2007}.


Motivated by these developments, we design a robotic visual inspection framework with perception-driven admittance control, such that force-based interaction control and perception-based orientation regulation are unified into a single compliant motion law for human-in-the-loop inspection. The framework combines a vision-based orientation controller that aligns the end-effector with estimated surface normals and an admittance controller that converts the resulting virtual forces into velocity commands for a collaborative manipulator. This enables stable and responsive robot--environment interaction while maintaining robustness to sensing noise and environmental uncertainty. While the focus of this work is robotic visual inspection, the proposed control framework is broadly applicable to other tasks involving perception-driven compliant orientation regulation, such as surface following, polishing, or contact-based manipulation.

\section{Proposed Framework}
\label{sec:framework}
\subsection{System Architecture and Control Pipeline}
\label{sec:architecture}
\begin{figure}[t]
    \centering
    \includegraphics[width=\linewidth]{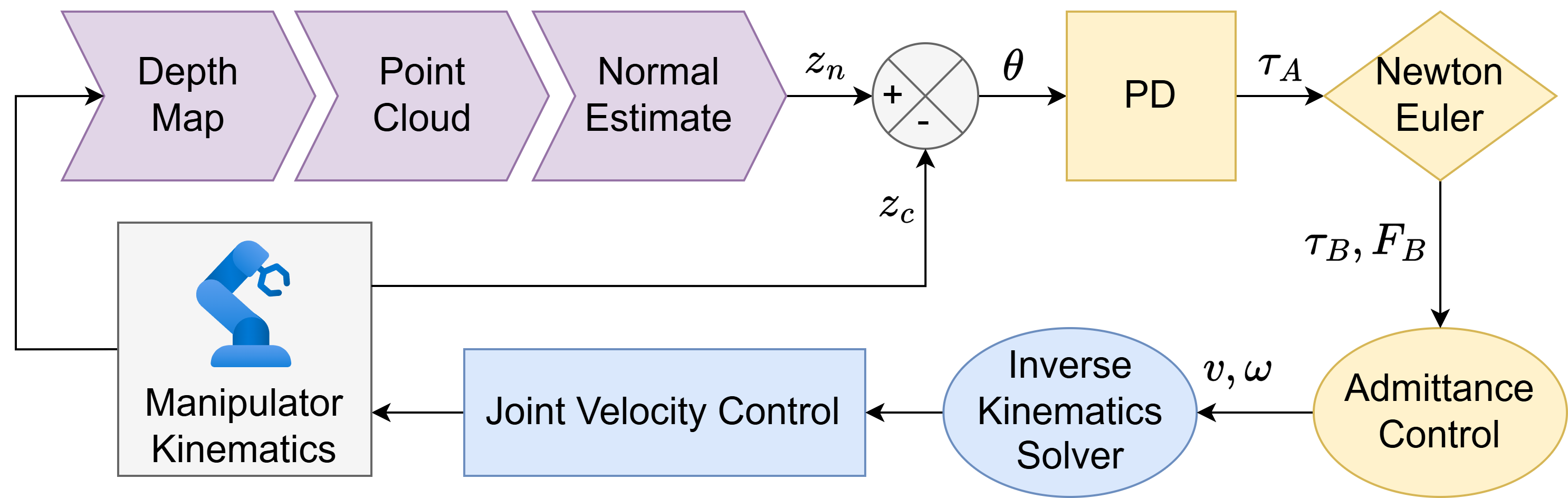}
    \caption{System architecture of the proposed perception-driven orientation control framework.}
    \label{fig:architecture}
\end{figure}
We design the overall architecture as a modular pipeline integrating perception, control, and motion execution, as illustrated in Fig.~\ref{fig:architecture}, such that perception-driven orientation regulation runs as an outer loop over the manufacturer's existing position/velocity controller without modification to the underlying servo architecture. The system is implemented on a UR5e industrial manipulator equipped with an eye-in-hand Intel RealSense D405 depth camera mounted at the end-effector. The camera provides high-resolution depth measurements of the local inspection surface, enabling real-time estimation of surface geometry.

The perception pipeline begins by acquiring depth images from the RealSense D405 camera. These depth images are converted into point clouds using the camera intrinsic calibration parameters. From the resulting point cloud, a local neighborhood corresponding to the inspection region is extracted and used to estimate the surface normal of the observed surface.

Surface normal estimation is performed using a combination of   RANSAC-based plane fitting \cite{fischler1981random} and principal component analysis (PCA) \cite{winn2023model}. First, the Random Sample Consensus (RANSAC) algorithm is used to robustly identify planar inliers within the point cloud while rejecting outliers caused by sensor noise or spurious depth measurements. This step ensures that the normal estimation process is not significantly affected by outliers or irregular surface points.

Once a consistent set of inlier points is identified, PCA is applied to compute the local surface normal. Given a set of $N$ points $\{\mathbf{p}_i\}$ belonging to the detected plane, the centroid of the point set is computed as
\begin{equation}
\mathbf{p}_c = \frac{1}{N} \sum_{i=1}^{N} \mathbf{p}_i.
\end{equation}
The covariance matrix of the centered points is then constructed as
\begin{equation}
C = \frac{1}{N} \sum_{i=1}^{N} (\mathbf{p}_i - \mathbf{p}_c)(\mathbf{p}_i - \mathbf{p}_c)^T.
\end{equation}
The eigenvector corresponding to the smallest eigenvalue of the covariance matrix represents the estimated surface normal direction. This normal vector defines the desired orientation for the robot end-effector during inspection.

The estimated surface normal is then compared with the current end-effector orientation to compute the orientation error. 

The orientation error is processed by the outer-loop controller, implemented as a PD controller that generates a virtual control torque. This torque drives the virtual sphere dynamic model described in Section~\ref{sec:orientation}, producing angular acceleration that reflects the desired correction motion.

The resulting angular acceleration is subsequently used to compute torque and force exerted at the end effector using the Newton--Euler equations. These force and torque are passed into the admittance model and are converted into Cartesian velocity commands in the robot task space.

The velocity commands are transmitted to the robot through the ROS~2 Servo node, which acts as the execution layer of the control architecture. The Servo node converts the task-space velocity commands into joint-level commands using the robot Jacobian and forwards them to the robot's internal controllers.

Because the ROS~2 Servo interface operates on top of the robot's existing joint servo loops, the proposed controller functions as an outer-loop compliance layer. This design allows perception-driven orientation regulation to be implemented without modifying the underlying robot control architecture, making the framework compatible with standard industrial robotic platforms.
\subsection{Admittance-Based Control Framework}
\label{sec:admittance}

We design the proposed control architecture with an admittance-based formulation, such that force and torque inputs are transformed into smooth, physically interpretable velocity commands for the robot end-effector while respecting actuator and servo constraints.

\begin{figure}[t]
\centering
\includegraphics[width=0.45\linewidth]{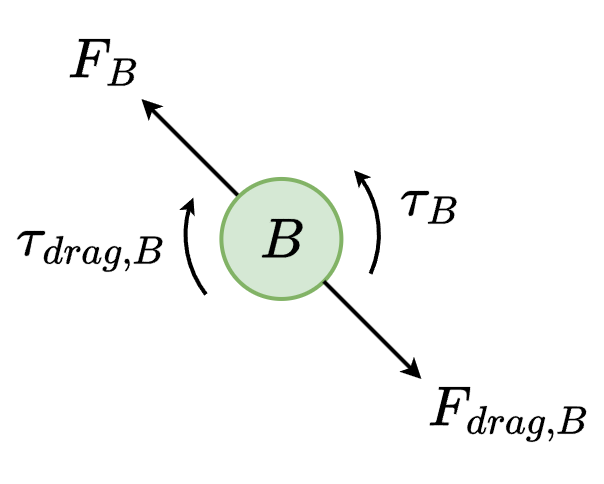}
\caption{Virtual mass--damper representation of the end-effector.}
\label{fig:virtual_mass_damper}
\end{figure}

To provide physical intuition, we design the end-effector as a virtual sphere of mass $m$ and radius $R$ moving in a viscous medium characterized by dynamic viscosity $\mu$, as illustrated in Fig.~\ref{fig:virtual_mass_damper}. This makes the resulting dynamics directly interpretable as a mass--damper system whose parameters carry concrete physical meaning. The inertia of the sphere about its center of mass at point $B$ is given by
\begin{equation}
I_B = \frac{2}{5} m R^2.
\end{equation}
The objective of this abstraction is not to reproduce the exact rigid-body
dynamics of the manipulator, but rather to construct a virtual
dynamic model that can be used to generate compliant motion
commands for orientation and teleoperation regulation.
The viscous interaction between the virtual sphere and the surrounding medium is modeled using translational and rotational drag forces. The translational drag force acting on the sphere is given by
\begin{equation}
F_{\mathrm{drag},B} = -c_d v_B,
\end{equation}
Similarly, the rotational drag torque is modeled as
\begin{equation}
\tau_{\mathrm{drag},B} = -\gamma \omega_B.
\end{equation}

Using these definitions, the net inputs to the admittance model are expressed as
\begin{equation}
F_{\mathrm{net}} = F_B - c_d v_B
\end{equation}
\begin{equation}
\tau_{\mathrm{net}} = \tau_B - \gamma \omega_B.
\end{equation}
The translational and rotational admittance dynamics are defined as
\begin{equation}
m \dot{v}_B = F_{\mathrm{net}}
\end{equation}
\begin{equation}
I_B \dot{\omega}_B = \tau_{\mathrm{net}}.
\end{equation}
The resulting accelerations are integrated to obtain the commanded velocities
\begin{equation}
v_B = \int \dot{v}_B \, dt, 
\qquad 
\omega_B = \int \dot{\omega}_B \, dt.
\end{equation}
These velocities are transmitted to the robot through the ROS~2 Servo node, allowing the orientation controller to act as a compliant outer-loop layer.
\subsection{Orientation Control}
\label{sec:orientation}

To generate the force and torque inputs required by the admittance model, we introduce a virtual rigid-body interpretation of the end-effector and pivot point. The goal of this model is not to capture the full manipulator dynamics, but to provide a physically consistent mapping from orientation error to equivalent forces and torques.

\begin{figure}[t]
\centering
\includegraphics[width=0.7\linewidth]{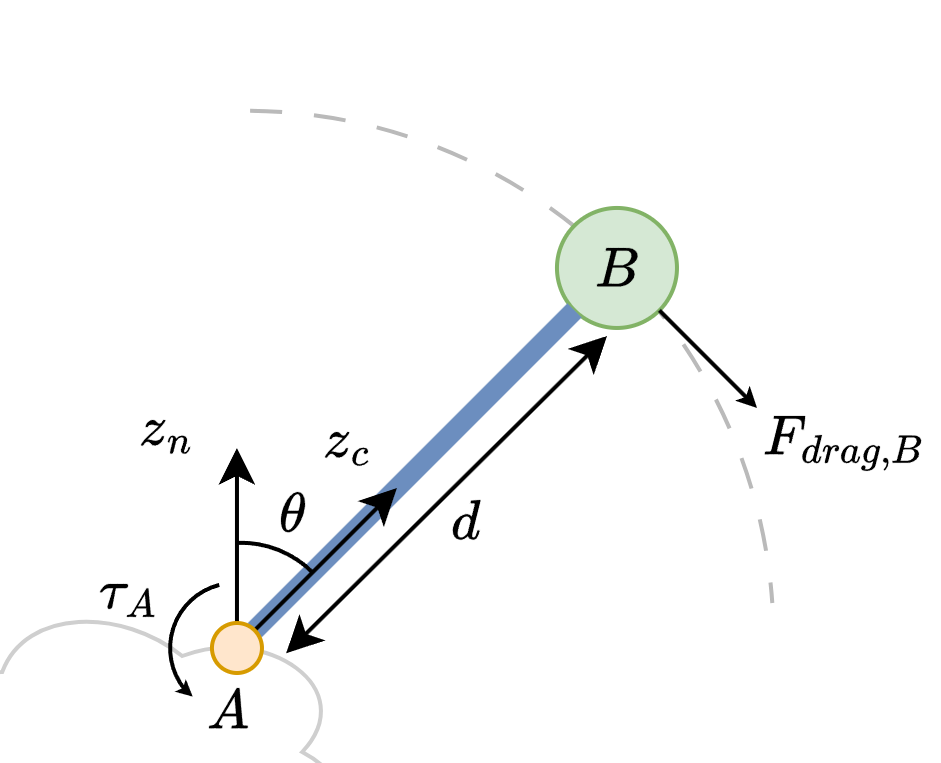}
\caption{Rigid-body interpretation of the virtual sphere.}
\label{fig:fbd_AB}
\end{figure}

As illustrated in Fig.~\ref{fig:fbd_AB}, let point $A$ denote the pivot associated with the local surface-normal frame, and let point $B$ denote the center of the virtual sphere located at a distance $d$ from point $A$. The control torque $\tau_A$ is applied at point $A$, which induces rotational motion of the virtual sphere.

\subsubsection{Rotational dynamics about pivot}

The applied torque $\tau_A$ produces angular acceleration $\alpha = \ddot{\theta}$ about point $A$. The inertia about the pivot is obtained using the parallel-axis theorem:
\begin{equation}
I_A = I_B + md^2,
\end{equation}
where $I_B = \frac{2}{5} m R^2$ is the inertia of the virtual sphere about its center of mass.

The rotational dynamics about point $A$ are given by
\begin{equation}
\tau_A = I_A \alpha - \tau_{\mathrm{drag}}.
\end{equation}
\subsubsection{Viscous drag torque}

The translational drag force acting on the sphere is modeled using Stokes' law:
\begin{equation}
F_{\mathrm{drag},B} = -6\pi \mu R v_B,
\end{equation}
where $v_B$ is the linear velocity of point $B$. Since $v_B = d\dot{\theta}$, the drag torque about point $A$ becomes
\begin{equation}
\tau_{\mathrm{drag}} = F_{\mathrm{drag},B} d = -6\pi \mu R d^2 \dot{\theta}.
\end{equation}
Substituting into the rotational dynamics yields
\begin{equation}
I_A \alpha + 6\pi \mu R d^2 \dot{\theta} = \tau_A.
\end{equation}
Solving for angular acceleration,
\begin{equation}
\alpha = \frac{\tau_A - 6\pi \mu R d^2 \omega_A}{I_A}.
\end{equation}
\subsubsection{Equivalent mass--damper dynamics}

The rotational dynamics can be expressed in the standard mass--damper form as
\begin{equation}
I_A \ddot{\theta} + b \dot{\theta} = \tau_A,
\label{eq:rot_dyn_compact}
\end{equation}
where the damping coefficient is defined as
\begin{equation}
b = 6\pi \mu R d^2.
\end{equation}
This representation highlights the physical interpretation of the virtual system: the inertia term $I_A$ resists angular acceleration, while the viscous term $b \dot{\theta}$ provides passive damping and ensures smooth motion.








\subsection{PD-Based Orientation Control}
\label{sec:pd}

Using the mass-damper dynamics from Section~\ref{sec:orientation}, we design a feedback control law to regulate the end-effector orientation, such that the closed-loop orientation error follows critically damped second-order dynamics while respecting actuator torque limits.

Let $\theta_{\mathrm{ref}}$ denote the desired orientation and define the tracking error as
\begin{equation}
e = \theta_{\mathrm{ref}} - \theta.
\end{equation}
A proportional-derivative (PD) control law is employed to generate the control torque applied at point $A$:
\begin{equation}
\tau_A = k_1 e + k_2 \dot{e}.
\label{eq:pd_control}
\end{equation}
\subsubsection{Closed-loop dynamics}

Substituting the control law \eqref{eq:pd_control} into the rotational dynamics in \eqref{eq:rot_dyn_compact} yields
\begin{equation}
I_A \ddot{\theta} + b \dot{\theta} = k_1 (\theta_{\mathrm{ref}} - \theta) + k_2 (\dot{\theta}_{\mathrm{ref}} - \dot{\theta}).
\end{equation}
For the regulation problem, $\theta_{\mathrm{ref}} = 0$ and $\dot{\theta}_{\mathrm{ref}} = 0$, which give
\begin{equation}
I_A \ddot{\theta} + (b + k_2)\dot{\theta} + k_1 \theta = 0.
\end{equation}
Thus, the characteristic polynomial of the closed-loop system is
\begin{equation}
s^2 + \frac{b + k_2}{I_A}s + \frac{k_1}{I_A} = 0.
\end{equation}
\subsubsection{Gain selection via second-order design}

To obtain a desired transient response, the closed-loop characteristic polynomial is matched with the standard second-order form
\begin{equation}
s^2 + 2\zeta \omega_n s + \omega_n^2 = 0,
\end{equation}
which yields the gain relations
\begin{equation}
k_1 = I_A \omega_n^2
\end{equation}
\begin{equation}
k_2 = 2\zeta \omega_n I_A - b.
\end{equation}
\subsubsection{Torque saturation constraint}

In practical implementation, the control torque must respect actuator limits. Under terminal velocity conditions, the viscous drag torque balances the applied torque, yielding the bound
\begin{equation}
\tau_{\max} = c_d d\, v_{\max},
\end{equation}
where
\begin{equation}
c_d = 6\pi \mu R.
\end{equation}
Thus, the applied torque is saturated as
\begin{equation}
\tau_{\mathrm{applied}} = \mathrm{sat}(\tau_A),
\qquad
|\tau_{\mathrm{applied}}| \le \tau_{\max}.
\end{equation}
\subsubsection{Design constraint from torque saturation}

In addition to enforcing actuator limits, the torque saturation condition provides a guideline for selecting controller gains.

For an initial condition $x_0 = [\theta_0,\,0]^T$, the PD control law in \eqref{eq:pd_control} gives the initial torque
\begin{equation}
\tau(0) = k_1 (\theta_{\mathrm{ref}} - \theta_0).
\end{equation}
For the regulation problem with $\theta_{\mathrm{ref}} = 0$, this becomes
\begin{equation}
\tau(0) = -k_1 \theta_0,
\qquad
|\tau(0)| = k_1 |\theta_0|.
\end{equation}
Using the gain relation $k_1 = I_A \omega_n^2$, the initial torque can be written as
\begin{equation}
|\tau(0)| = I_A \omega_n^2 |\theta_0|.
\end{equation}
To ensure that the actuator limits are not exceeded for the maximum expected orientation error $\theta_{\max}$, the following condition must hold:
\begin{equation}
I_A \omega_n^2 \theta_{\max} \le \tau_{\max}.
\end{equation}
Solving for the allowable natural frequency yields
\begin{equation}
\omega_{n,\max} = \sqrt{\frac{\tau_{\max}}{I_A \theta_{\max}}}.
\end{equation}
To provide robustness against modeling uncertainty and unmodeled dynamics, a safety factor $s \in (0,1)$ is introduced:
\begin{equation}
\omega_n = s \sqrt{\frac{\tau_{\max}}{I_A \theta_{\max}}}.
\end{equation}
\subsubsection{Critically damped response}

To ensure fast convergence without overshoot, the critically damped case is selected by choosing
\begin{equation}
\zeta = 1.
\end{equation}
Under this condition, the closed-loop poles are
\begin{equation}
p_1 = p_2 = -\omega_n,
\end{equation}
resulting in a monotonic response with the fastest possible settling time without oscillations. This behavior is desirable for orientation regulation tasks where overshoot may lead to misalignment with the surface.
\subsubsection{Connection to admittance dynamics}

The torque $\tau_A$ generated by the PD controller determines the angular acceleration $\alpha$ through the rotational dynamics. This angular motion is subsequently used to compute the equivalent force and torque inputs to the admittance model, which generates the commanded translational and rotational velocities for the robot end-effector.
\subsection{Newton--Euler Dynamics and Net Inputs to the Admittance Model}
\label{sec:newton_euler}

\begin{figure}[t]
\centering
\includegraphics[width=0.7\linewidth]{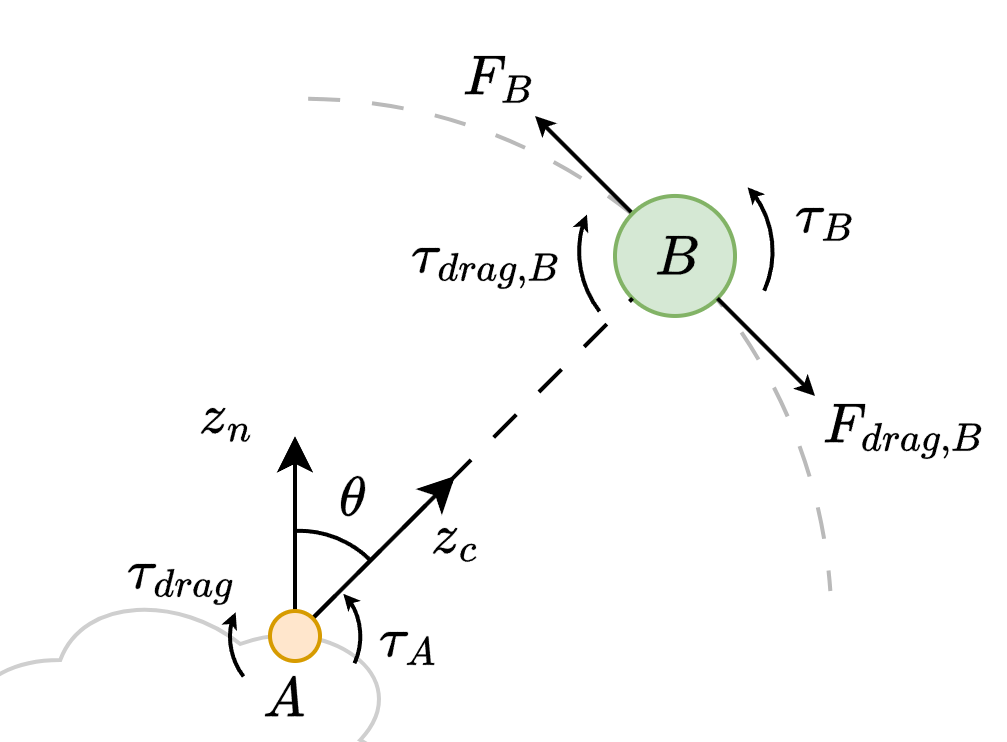}
\caption{Free-body representation of the virtual sphere used in the admittance model.}
\label{fig:virtual_sphere_dynamics}
\end{figure}

After $\alpha$ is obtained, we reinterpret the virtual rigid-body as an equivalent free body floating in a viscous medium, as illustrated in Fig.~\ref{fig:virtual_sphere_dynamics}. In this free-body view, point $B$ represents the center of mass of the virtual sphere, and the equivalent force and torque acting on the sphere are used to construct the inputs to the admittance model. Thus, the rigid-body model is used only to determine the angular acceleration, whereas the free-body representation is used to determine the net force and torque driving the admittance dynamics.

\subsubsection{Linear dynamics of the virtual sphere}

Once the angular acceleration $\alpha$ is known, the linear acceleration of the sphere center is obtained as
\begin{equation}
a_B = \alpha d,
\end{equation}
where $d$ is the distance between points $A$ and $B$.

Applying Newton's second law to the free body at point $B$ gives
\begin{equation}
F_B = m(\alpha d) + 6\pi \mu R v_B.
\end{equation}
\subsubsection{Rotational dynamics of the virtual sphere}

Similarly, applying Euler's rotational equation about the center of mass gives
\begin{equation}
\tau_B = I_B \alpha + \gamma \omega_B,
\end{equation}
where $I_B = \frac{2}{5}mR^2$ and $\omega_B = \omega_A$.

In order to preserve a 1:1 spin-orbit resonance when dropping the rigid body model, a modified version of Stokes' law is applied to the sphere:
\begin{equation}
\gamma = 2.4\pi \mu R^3.
\end{equation}
This value is approximately $3.3$ times smaller than the classical Stokes rotational drag coefficient ($8\pi\mu R^3$).

\subsubsection{Net inputs to the admittance model}

The net force and torque supplied to the admittance model are obtained by subtracting the viscous drag terms from the equivalent free-body quantities:
\begin{equation}
F_{\mathrm{net}} = F_B - 6\pi \mu R v_B
\end{equation}
\begin{equation}
\tau_{\mathrm{net}} = \tau_B - \gamma \omega_B.
\end{equation}
These effective inputs are then used in the admittance dynamics of Section~\ref{sec:admittance} to generate the commanded translational and rotational velocities of the end-effector.

\section{Experimental Validation}
\label{sec:experiments}

To validate the proposed orientation control and admittance framework, experiments were conducted using the UR5e manipulator equipped with an eye-in-hand Intel RealSense D405 depth camera. The objective of the experiment is to regulate the end-effector orientation by aligning the camera optical axis with the estimated surface normal obtained from the point cloud.

The experiments were performed under two different initial error conditions in order to evaluate the behavior of the controller both within and beyond the torque saturation region. The experimental results were compared with a synthetic MATLAB simulation that emulates the same control pipeline, including the orientation PD controller and the admittance dynamics.

In addition, a baseline comparison was conducted using manual teleoperation. A total of 10 trials were performed in which a human operator manually oriented the end-effector to align with the target surface without any perception-driven or automated assistance. This comparison provides insight into the repeatability and efficiency of the proposed perception-driven control framework relative to human operation.

\begin{figure}[t]
\centering
\includegraphics[width=\linewidth]{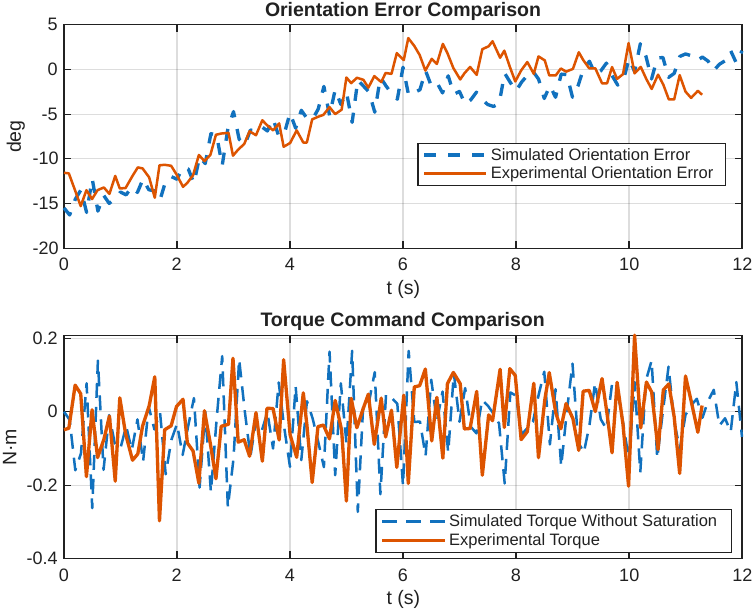}
\caption{Comparison of orientation error and torque command between simulation and experiment for the $15^\circ$ initial orientation error case.}
\label{fig:angle_torque_15}
\end{figure}
\begin{figure}[t]
\centering
\includegraphics[width=\linewidth]{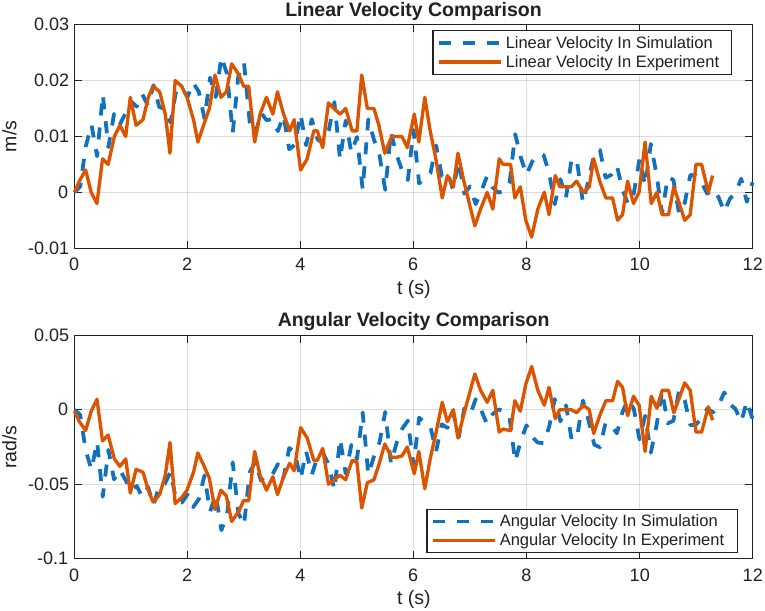}
\caption{Comparison of linear and angular velocities between simulation and experiment for the $15^\circ$ initial orientation error case.}
\label{fig:vel15}
\end{figure}
\begin{figure}[t]
\centering
\includegraphics[width=\linewidth]{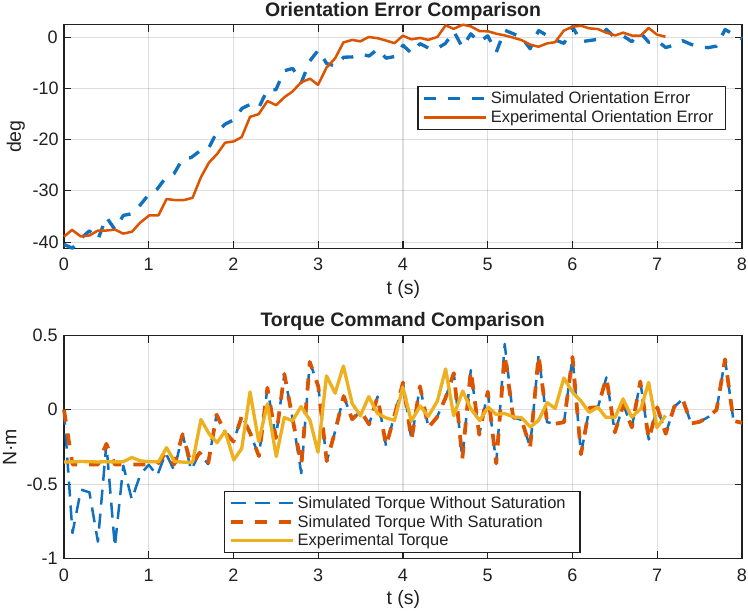}
\caption{Comparison of orientation error and torque command between simulation and experiment for the $40^\circ$ initial orientation error case.}
\label{fig:angle_torque_40}
\end{figure}
\begin{figure}[t]
\centering
\includegraphics[width=\linewidth]{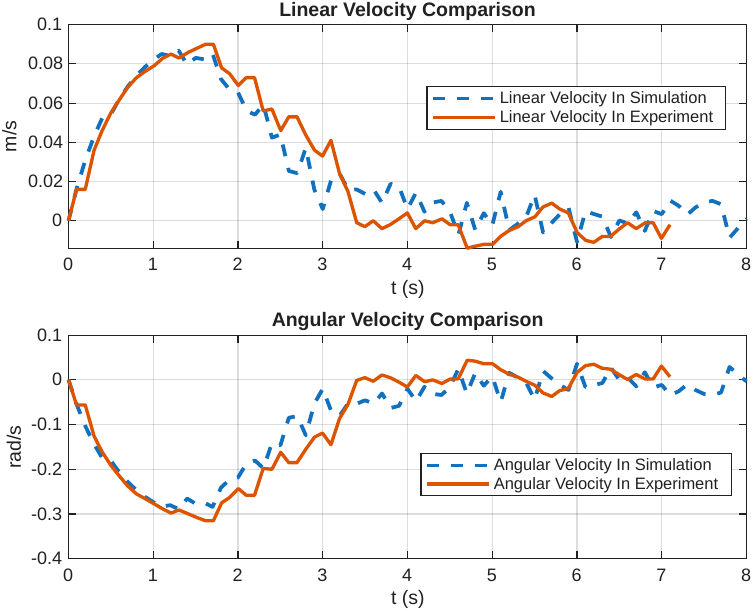}
\caption{Linear and angular velocity comparison between simulation and experiment for the $40^\circ$ initial orientation error case.}
\label{fig:vel40}
\end{figure}

\subsection{Experimental Conditions}

\subsubsection{Nominal Parameter Selection}

Based on the parameter sweeps, the nominal admittance parameters were selected as
\[
m = 2.5\,\mathrm{kg}, \qquad
\mu = 1.0, \qquad
R = 0.65\,\mathrm{m}.
\]
These values provide a balanced response: the chosen mass is a trade-off between speed of convergence and aggression of motion, the viscosity prevents excessive torque-driven acceleration, and the radius balances the responsiveness that might have been affected by the value of viscosity. Overall, this parameter set was selected to achieve stable convergence, moderate velocity magnitudes, and smooth motion suitable for perception-driven robotic inspection.

\subsubsection{Small Initial Error Case}
In the first experiment, the initial error was set to $15^\circ$ with the
saturation threshold set to $\theta_{\max} = 90^\circ$, ensuring that the
controller operated entirely in the unsaturated regime. The maximum linear
velocity and camera offset from the pivot were fixed at
$v_{\max} = 0.1\,\mathrm{m/s}$ and $d = 0.3\,\mathrm{m}$, respectively, and
held constant across both experiments. Under these conditions, the controller
operates without torque saturation and the system response is dominated by the
nominal admittance dynamics. Results are shown in Figs.~\ref{fig:angle_torque_15} and \ref{fig:vel15}.

\subsubsection{Large Initial Error Case}
In the second experiment, the initial error was increased to $40^\circ$ and the
saturation threshold was lowered to $\theta_{\max} = 20^\circ$, causing the
controller to enter the saturated regime during the initial phase of motion.
All other parameters were unchanged from the previous case. When the initial
orientation error exceeds $\theta_{\max}$, the controller torque saturates at
the actuator limit, and the end-effector initially moves close to the maximum
allowable velocity before transitioning to nominal admittance-controlled
behavior as the orientation error decreases. Results are shown in Figs.~\ref{fig:angle_torque_40} and \ref{fig:vel40}.

For each of the experiments, a final, mean orientation error was computed using 10 samples at the end of the trial. Repeatably, a value of $0.4^\circ$ was obtained.
\section{Discussion}
\label{sec:discussion}

The experimental results demonstrate that the proposed admittance-based orientation control framework produces stable and consistent behavior across different initial conditions. In all tested cases, the orientation error converges toward zero while maintaining smooth velocity and torque profiles. This confirms that the virtual sphere dynamics combined with the PD orientation controller provide an effective mechanism for regulating end-effector orientation during robotic inspection tasks.

A key observation from the results is the close agreement between the synthetic trajectories obtained from the admittance-integrated simulation and the measurements recorded from the robot. The angular error, angular velocity, linear velocity, and torque command signals exhibit similar trends in both the simulation and hardware data, as illustrated in Figs.~\ref{fig:angle_torque_15}, \ref{fig:vel15}, \ref{fig:angle_torque_40}, and \ref{fig:vel40}. Minor discrepancies are primarily attributed to sensor noise, uncertainty in surface normal estimation, and the internal dynamics of the robot controller. Nevertheless, the overall agreement indicates that the simplified mass–damper representation used in the admittance model captures the dominant behavior of the physical system.

An additional behavior is observed in the experiment corresponding to the larger initial angle. When the initial orientation error exceeds the design limit $\theta_{\max}$ used in the torque constraint formulation, the commanded torque reaches the saturation bound $\tau_{\max}$, as seen in Fig.~\ref{fig:angle_torque_40}. Under this condition, the controller operates in a torque-limited regime where the applied torque is clipped at the actuator limit. As a result, the end-effector initially moves at the maximum admissible velocity $v_{\max}$ determined by the actuator constraints.

This behavior is further reflected in the velocity profiles shown in Fig.~\ref{fig:vel40}, where the linear velocity reaches the imposed $v_{\max}$ bound during the initial phase of motion before decreasing as the orientation error falls within the controllable range. This observation is consistent with the controller design presented earlier, where the torque saturation constraint was used to derive the maximum allowable system response.
In addition to the perception-driven experiments, a set of manual teleoperation trials was conducted to evaluate human performance as a baseline. As shown in Table \ref{tab:manual_reorientation}, a total of 10 trials were performed in which an operator manually oriented the end-effector from an initial offset of $30^\circ$ until they judged the camera principal axis to be parallel to the surface normal. This resulted in a mean completion time of 8s with a variability of 1.5s compared to our algorithm's repeatable convergence of 6s. This indicates limited repeatability in manual operation, as the performance depends on the operator's skill, reaction time, and subjective judgment, leading to inconsistent task execution across trials.

\begin{table}[t]
\centering
\caption{Reorientation time across manual trials.}
\label{tab:manual_reorientation}
\begin{tabular}{cc}
\toprule
Manual Trial & Reorientation Time (s) \\
\midrule
1  & 10.5 \\
2  & 6.5  \\
3  & 6.2  \\
4  & 8.4  \\
5  & 7.5  \\
6  & 8.2  \\
7  & 10.0 \\
8  & 8.2  \\
9  & 6.0  \\
10 & 8.5  \\
\bottomrule
\end{tabular}
\end{table}

In contrast, the proposed perception-driven orientation control framework exhibited highly consistent behavior across repeated runs. The automated controller produced nearly identical convergence profiles and task completion times for similar initial conditions, demonstrating strong repeatability and robustness. This highlights a key advantage of the proposed method over manual operation, namely its ability to deliver predictable and repeatable performance independent of human variability.

A comparison with existing adaptive orientation control approaches further highlights the effectiveness of the proposed framework. In the work by Nakhaeinia \cite{nakhaeinia2018hybrid}, they achieved a final mean orientation error of $0.4^\circ$ under position-based iterative adaptive control utilizing IR sensing. Our proposed perception-driven admittance control framework achieves equivalent performance while incorporating real-time control, human-in-the-loop compliance, and a force-based adaptive control strategy.

Overall, the controller demonstrates robustness to perception noise and varying initial orientation errors and validates the proposed admittance-based control framework, indicating that the MATLAB simulation provides a reliable representation of the real robot dynamics.

\section{Future Work}
\label{sec:future}

An important direction for future work is extending the proposed framework to support the fusion of multiple control objectives within a unified dynamic structure. Admittance control provides a natural interface for such integration, as it maps generalized wrench or interaction objectives into motion commands.

Through this formulation, perception-driven orientation corrections, operator teleoperation inputs, and disturbance rejection mechanisms can be expressed as equivalent inputs to the same admittance dynamics. Additional sensing objectives, such as autofocus control, can also be incorporated within this framework. This enables multiple control objectives to be combined without modifying the underlying robot servo architecture.
Incorporation of a kinematic Kalman filter or depth fusion will help reduce the final mean orientation error further.
A promising extension of this work is the development of an operator-assisted inspection system where teleoperation inputs provide coarse positioning while the perception-driven controller performs automatic orientation alignment using estimated surface normals. Future research will also investigate the use of PID-based controllers within this framework to reduce tracking error during teleoperation and regulate interaction dynamics when multiple control objectives are present, enabling more flexible and adaptive robotic inspection.

\section{Conclusion}
\label{sec:conclusion}

Precision visual inspection is a critical determinant of yield, cost, and safety across aerospace, semiconductor, and medical manufacturing, yet existing automation either pre-plans the inspection trajectory offline or excludes the human operator from the control loop. To address this gap, this paper presented a novel application of real-time, closed-loop robotic orientation control to precision visual inspection: an admittance-based surface alignment framework for human-in-the-loop operation. We designed the end-effector as a virtual sphere in a viscous medium, such that the resulting physically interpretable mass--damper system unifies perception-driven orientation correction and operator input into a single compliant motion command stream. Surface normals estimated from real-time depth point clouds provide continuous orientation feedback, while a PD controller generates virtual control torques that drive the admittance dynamics. The resulting velocity commands are executed through the robot's existing servo architecture via inverse kinematics, allowing the controller to operate as a modular outer-loop compliance layer without modification to the underlying position-controlled platform.
Experimental validation on a UR5e manipulator with an eye-in-hand Intel RealSense D405 camera demonstrates stable orientation regulation under both unsaturated and torque-saturated conditions, with strong agreement between the admittance model predictions and hardware response. The framework achieves a final mean orientation error of $0.4^\circ$, matching the performance of prior position-based iterative methods that rely on infrared sensing, while additionally providing real-time closed-loop control and a force-based formulation naturally suited to human-in-the-loop compliance. Baseline comparison with manual teleoperation further highlights the repeatability and consistency of the proposed approach relative to unassisted human operation.
The proposed framework therefore establishes a practical foundation for integrating perception-driven orientation regulation with shared autonomy in robotic inspection, enabling human operators to focus on high-level classification and decision-making while the controller maintains surface alignment automatically.


%








\bibliographystyle{IEEEtran} 
 \bibliography{references} 
\end{document}